\documentclass[conference]{IEEEtran}
\IEEEoverridecommandlockouts
\usepackage{cite}
\usepackage{amsmath,amssymb,amsfonts}
\usepackage{graphicx}
\usepackage{textcomp}
\usepackage{xcolor}
\usepackage{algorithm}
\usepackage{algorithmicx}

\usepackage{marvosym} 
\usepackage{url}
\usepackage{bm}
\usepackage{algpseudocode}   
\usepackage{multirow}
\usepackage{diagbox}

\usepackage{subfigure}

\begin{document}

\title{MODRL/D-EL: Multiobjective Deep Reinforcement Learning with Evolutionary Learning for Multiobjective Optimization\\

\thanks{This work is supported by the National Key R\&D Program of China (2018AAA0101203), the National Natural Science Foundation of China (62072483), and the Natural Science Foundation of Guangdong Province (2018A030313703, 2021A1515012298). (\it{Corresponding author: Jiahai Wang.})}
}

\author{\IEEEauthorblockN{1\textsuperscript{st} Yongxin Zhang}
\IEEEauthorblockA{\textit{School of Computer Science and Engineering} \\
\textit{Sun Yat-sen University}\\
Guangzhou, P. R. China \\
zhangyx266@mail2.sysu.edu.cn}
\\
\IEEEauthorblockN{3\textsuperscript{rd} Zizhen Zhang}
\IEEEauthorblockA{\textit{School of Computer Science and Engineering} \\
	\textit{Sun Yat-sen University}\\
	Guangzhou, P. R. China \\
	zhangzzh7@mail.sysu.edu.cn}
\and
\IEEEauthorblockN{2\textsuperscript{nd} Jiahai Wang\textsuperscript{\Letter} }
\IEEEauthorblockA{\textit{School of Computer Science and Engineering} \\
\textit{Sun Yat-sen University}\\
Guangzhou, P. R. China \\
wangjiah@mail.sysu.edu.cn}

\\
\IEEEauthorblockN{4\textsuperscript{th} Yalan Zhou}
\IEEEauthorblockA{\textit{Colloge of Information} \\
\textit{Guangdong University of Finance and Economics}\\
Guangzhou, P. R. China \\
zhouylan@163.com}
}

\maketitle

\begin{abstract}
Learning-based heuristics for solving combinatorial optimization problems has recently attracted much academic attention. While most of the existing works only consider the single objective problem with simple constraints, many real-world problems have the multiobjective perspective and contain a rich set of constraints. This paper proposes a multiobjective deep reinforcement learning with evolutionary learning algorithm for a typical complex problem called the multiobjective vehicle routing problem with time windows (MO-VRPTW). In the proposed algorithm, the decomposition strategy is applied to generate subproblems for a set of attention models. The comprehensive context information is introduced to further enhance the attention models. The evolutionary learning is also employed to fine-tune the parameters of the models. The experimental results on MO-VRPTW instances demonstrate the superiority of the proposed algorithm over other learning-based and iterative-based approaches.
\end{abstract}

\section{Introduction}
A multiobjective optimization problem (MOP) can be generally defined as follows,
\begin{equation}
	\begin{split}
		&\mathrm{Min}\quad F(x)=(f_1(x),f_2(x),...,f_m(x))^\mathrm{T},\\
		&s.t.\quad x\in X,
	\end{split}
\end{equation}
where $X$ is the solution space, $F(x)$ is a $m$-dimensional vector in which $f_i$ is the $i$-th objective function of the MOP. Since these $m$ objective functions potentially conflict with each other, one would be keen in finding a set of trade-off solutions among which the most desirable solution can be chosen.

Formally, let $x,y\in \mathbb{R}^m$ be two objective vectors of MOP. $x$ is said to dominant to $y$, if $x_i\leq y_i,\ \forall i \in \{1,2,...,m\}\ \mathrm{and} \ x_i<y_i,\ \exists i \in \{1,2,...,m\} $, denoted as $x\prec y$. A solution $x^*$ is called Pareto optimal solution, if $F(x)\not \prec F(x^*),\ \forall x\in X$. The set which contains all the Pareto optimal solutions is called the Pareto set (PS), and the set $\{F(x)|x\in \mathrm{PS}\}$ is called the Pareto frontier (PF) \cite{zhang2007moea}.

In literature, many iterative-based algorithms are proposed to solve MOPs. These algorithms can be roughly divided into two categories: the multiobjective evolutionary algorithms (MOEAs) \cite{zhou2011multiobjective} (e.g., MOEA/D \cite{zhang2007moea} and NSGA-\uppercase\expandafter{\romannumeral2} \cite{deb2002a}), and multiobjective local search based algorithms (e.g., MOGLS \cite{jaszkiewicz2002genetic}). MOEAs search for PF by evolving a population of solutions simultaneously. MOGLS uses local search operators to improve solutions iteratively. These algorithms perform well in many multiobjective benchmark problems. However, as presented in \cite{li2020deep}, the iterative-based algorithms have some drawbacks. First, to find near-optimal solutions, a large number of iterations are required, so it would be very time consuming. Second, the iteration process needs to be re-performed once a slight change of problem happens or a similar instance is encountered. Third, most heuristic operators used in these algorithms heavily rely on experts' empirical knowledge, which may affect the solution quality.

Recently, learning-based heuristics for solving optimization problems has attracted more and more attention \cite{Bengio2020,mazyavkina2020reinforcement,vesselinova2020learning}. In particular, deep reinforcement learning (DRL) has made remarkable achievement in solving single objective problems \cite{bello2017neural,dai2017learning,nazari2018reinforcement,deudon2018learning,kool2019attention}. In addition, based on the decomposition strategy \cite{zhang2007moea} and DRL algorithm, a multiobjective deep reinforcement learning (MODRL) framework \cite{li2020deep} is proposed for MOPs. Following this framework, \cite{wu2019modrl} proposes a multiobjective deep reinforcement learning algorithm with attention model (MODRL/D-AM), which achieves good performance on multiobjective traveling salesman problem. The characteristics of these works are summarized in Table I.
\begin{table*}[]
	\label{related works}
		\caption{The characteristics of related works and our paper}
	
	\begin{tabular}{|c|c|c|c|l|c|}
		\hline
		\textbf{Objective}                   & \textbf{Literature} & \textbf{Problem} & \textbf{Encoder}       & \multicolumn{1}{c|}{\textbf{Context information}}                                                                           & \textbf{Training method} \\ \hline
		\multirow{5}{*}{Single objective optimization} & \cite{bello2017neural}            & TSP              & LSTM                   & last visited node                                                                                                           & DRL                      \\ \cline{2-6} 
		& \cite{dai2017learning}           & TSP              & GNN                    & last visited node                                                                                                           & DRL                      \\ \cline{2-6} 
		& \cite{nazari2018reinforcement}            & VRP              & linear embedding       & last visited node                                                                                                           & DRL                      \\ \cline{2-6} 
		& \cite{deudon2018learning}            & TSP              & attention model        & last $k$ visited nodes                                                                                                        & DRL                      \\ \cline{2-6} 
		& \cite{kool2019attention}            & TSP,VRP          & attention model        & \begin{tabular}[c]{@{}l@{}}graph information, last visited node, \\ first visited node (depot for VRP)\end{tabular}         & DRL                      \\ \hline
		\multirow{3}{*}{Multiobjective optimization}   & \cite{li2020deep}             & MO-TSP           & LSTM                   & last visited node                                                                                                           & DRL                      \\ \cline{2-6} 
		& \cite{wu2019modrl}            & MO-TSP           & attention model        & \begin{tabular}[c]{@{}l@{}}graph information, last visited node,\\ first visited node\end{tabular}                          & DRL                      \\ \cline{2-6} 
		& Our paper              & MO-VRPTW         & attention model + MLP & \begin{tabular}[c]{@{}l@{}}graph information, last visited node, \\ depot, fleet information, route information\end{tabular} & DRL + EL                \\ \hline
	\end{tabular}
\end{table*}

In this paper, we initiate the study on tackling an MOP with complex constraints, called multiobjective vehicle routing problem with time windows (MO-VRPTW). MO-VRPTW is more applicable in real-world scenarios. We try to apply learning-based heuristics for this problem. We have found that the attention model \cite{kool2019attention} in MODRL/D-AM constructs solution by sequentially adding nodes to partial solution. The selection of next node is based on very limited context information, which may lead to inefficiencies in MOPs (e.g., MO-VRPTW). Besides, the diversity of solutions obtained by MODRL/D-AM is not as good as expected \cite{li2020deep}. To address these issues, this paper proposes a multiobjective deep reinforcement learning and evolutionary learning algorithm (MODRL/D-EL). In the proposed algorithm, two mechanisms are employed. First, the comprehensive context information (CCI) of MOPs  is taken into consideration. The attention model with CCI can make more reasonable decision in constructing solution. Second, a hybrid training method composed of deep reinforcement learning and evolutionary learning (EL) is employed to train the models. 
The comparisons between our method and the related works are shown in Table I.

To summarize, the contributions of this paper are threefold:
\begin{itemize}
	\item The attention model with comprehensive context information is introduced to solve MO-VRPTW. It is shown to be effective in improving the solution quality.
	\item A two-stage hybrid learning strategy is presented. In the first stage, DRL is employed to train models. In the second stage, evolutionary learning is used to fine-tune the parameters of models. The results indicate that both convergence and diversity of solutions can be improved. 
	\item Computational experiments are conducted on both random instances and classical benchmark instances. It is demonstrated that the proposed method outperforms several existing methods.
\end{itemize}

The remaining of this paper is structured as follows. In Section II, problem formulation of MO-VRPTW is introduced. MODRL/D-AM is briefly reviewed in Section III. Thereafter, Section IV provides a detailed description of the proposed MODRL/D-EL. Experimental results are shown and analyzed in Section IV. Conclusions are drawn in Section V.

\section{Problem Formulation}

MO-VRPTW is a classical MOP with complex constraints and has a high relevance for many practical applications. It can be defined on a complete undirected graph $G=\{\mathcal{V},\mathcal{E}\}$. Node set $\mathcal{V}=\{v_i|i=0,1,...,N\}$, where node $i = 0$ is the depot and $i \in \{1, . . . , N\}$ are the
customers. Each node $v_i$ has some attributes including Euclidean coordinates $c_i \in \mathbb{R}^2$, a time window constraint $t_i\in \mathbb{R}^2$, and a demand $q_i \in \mathbb{R}^+$ that requires to be satisfied. The time window constraint $t_i$ indicates a tuple $[a_i,b_i],\ a_i\leq b_i$, where $a_i$ and $b_i$ are lower and upper bounds of the possible service time. Moreover, the service at each customer $v_i \in \mathcal{V}$ requires an identical service duration $w$. For the depot, $q_0=0$ and $t_0=[a_{0},b_{0}]$, where $a_{0}$ and $b_{0}$ denote the earliest possible departure from and latest return to the depot, respectively. $\mathcal{E}=\{e_{ij}|i,j\in \mathcal{V}\}$ is the edge set of graph $G$. The traveling distance of $e_{ij}$ is defined as the Euclidean distance between $v_i$ and $v_j$, and the traveling time of $e_{ij}$ is equal to the traveling distance numerically. There is a fleet contains $K$ uniform vehicles with identical capacity $Q>0$. Vehicle $k$ departures from the depot, serves customers with respect to capacity and time windows constraints, and finally returns to the depot. The sequence of customers that the vehicle severed constitutes a route $r_k$. A solution is a set of routes, denoted as $s=\{r_1,r_2,...,r_K\}$. A solution $s$ needs to meet the following constraints: (1) all customers are visited exactly once, i.e., $\bigcup_{r\in s}=\mathcal{V}$ and $r_a\bigcap r_b=\emptyset, \ \forall a,b\in \{1,2,...,K\}$; (2) all routes satisfy the capacity constraint, i.e., $\sum_{i\in r_k}q_i \leq Q$; (3) all customers are served within their specific time windows.

This paper focuses on two conflicting objectives, which are commonly adopted in previous work \cite{castro-gutierrez2011nature}: the total traveling distance and makespan, formally described as follows.
\begin{equation}
	\label{objectives}
	\begin{split}
		&\min\quad F(s)=(f_1(s),f_2(s))^\mathrm{T},\\
		&f_1(s)=\sum_{k=1}^{K}c_k,\\
		&f_2(s)=\max(c_1,c_2,...,c_K),\\
		&s.t.\  s\in S. 
	\end{split}
\end{equation}
where $c_k$ is the traveling distance of $k$-th route in solution $s$. $S$ is the valid solution space under the capacity and time windows constraints.

\section{Brief Review of MODRL/D-AM}

\subsection{Decomposition Strategy}\label{AA}

A well-known decomposition strategy \cite{zhang2007moea} is employed in MODRL/D-AM for MOPs. Specifically, given a set of uniformly spread weight vectors $\Lambda=\{{\lambda}_1,\lambda_2,...,\lambda_M\},\ \lambda_i=(\lambda_i^1,...,\lambda_i^m)^\mathrm{T}$, MODRL/D-AM adopts the weighted sum \cite{miettinen2012nonlinear} approach to convert MOP into $M$ single objective subproblems.
Then, a set of neural network models  $\mathrm{\Pi}=\{\pi_1,\pi_2,...,\pi_M\}$ is trained for tackling these subproblems, where $\pi_i$ corresponds to the subproblem with weight vector $\lambda_i$.

The MO-VRPTW is decomposed into $M$ single objective subproblems through decomposition strategy. 
For each subproblem, reinforcement learning is employed to train a neural network model to solve it. The states, actions, transition, rewards and policy in  the reinforcement learning framework are defined as follows:
\begin{itemize}
	\item {\it States}: a state consists of two parts. The first part includes the partial solution of subproblem and the candidate nodes. The second part is the position, remaining capacity, travel distance and travel time of each vehicle;

	\item {\it Actions}: an action is the next node to visit;
	\item {\it Transition}: transition corresponds to adding the selected action (node) into partial solution, and updating the state of corresponding vehicle;
	\item {\it Rewards}: The reward function of $i$-th subproblem is defined as the weighted sum of two objectives, i.e. $r_i=\lambda_i^{\mathrm{T}}F(s)$;
	\item {\it Policy}: policy is parameterized with a deep neural network model $\pi_i$ that selects a candidate node as next visiting node according to current state. The details of neural network model will be introduced in the following subsection.
\end{itemize}

\subsection{Model of subproblem: Attention Model}
In MODRL/D-AM, an attention model \cite{kool2019attention} is employed for a subproblem. The attention model, say $\pi$, is based on the encoder-decoder architecture, parametrized by $\boldsymbol{\theta}$.
\subsubsection{Encoder}
The encoder first takes the node-wise features $x_i=(c_i,t_i,q_i),  i\in \mathcal{V}$,  as input.
It computes the initial embedding $h_i^{(0)} \in \mathbb{R}^{d_{\mathrm{emb}}}$ ($d_{\mathrm{emb}}$ is set to 128 empirically) of $v_i$ through a linear projection, i.e., $h_i^{(0)}=Wx_i+B$. Then the initial embedding $h_i^{(0)}$ is updated through a stack of three self-attention (SA) blocks, i.e.,
\begin{equation}
	h_i=\mathrm{SA}(\mathrm{SA}(\mathrm{SA}(h_i^{(0)},H^{(0)})))\quad i\in \mathcal{V},
\end{equation}
where $H^{(0)}=(h_0^{(0)},h_1^{(0)},...,h_N^{(0)})$ is the sequence of initial embeddings. SA block consists of two sublayers: a multi-head attention layer (MHA) and a node-wise fully connected feed-forward layer (FF). Each sublayer also includes a skip-connection \cite{he2016deep} and batch normalization (BN) \cite{ioffe2015batch} i.e. 
\begin{equation}
	\hat h_i=\mathrm{BN}^{l}(h_i^{(l-1)}+\mathrm{MHA}^{l}(h_i^{(l-1)},H^{(l-1)})),
\end{equation}
\begin{equation}
	h_i^{(l)}=\mathrm{BN}^{l}(\hat h_i+\mathrm{FF}^{l}(\hat h_i)),
\end{equation}
where  $h^{(l)}_i$ denotes the node embedding produced by block $l$.
The multi-head attention layer is a linear combination of multiple (e.g., eight) single-head attention (SHA) layers, each of which takes a slice of all the input elements, formally defined as follows,
\begin{equation}
	\mathrm{MHA}(h,H;W)=\sum_{i=1}^ZW_i^{\mathrm{head}}\mathrm{SHA}(h_{\mathrm{slice}(i)},H_{\mathrm{.,slice}(i)};W),
\end{equation}
\begin{equation}
	\mathrm{SHA}(h,H;W)=\sum_{j=1}^{|H|}\mathrm{attn}(h,H;W^{\mathrm{query}},W^{\mathrm{key}})_jW^{\mathrm{value}}H_j,
\end{equation}
\begin{equation}
	\mathrm{attn}(h,H;W)=\mathrm{softmax}(\frac{1}{\sqrt{d_{key}}}h^\mathrm{T}(W^{\mathrm{query}})^\mathrm{T} W^{\mathrm{key}} H).
\end{equation}
The fully connected layer is defined as follows,

\begin{equation}
	\mathrm{FF}(h_i;W,B)=\mathrm{max}(0,Wh_i+B).
\end{equation}

\subsubsection{Decoder}
The decoder sequentially constructs solution in $T$ time steps. In each step, decoder outputs a node for the next visit. If the next visited node is depot, the current vehicle route is closed and a new vehicle is dispatched until all the customers have been served. Specifically, at each time step $t \in \{1,2,...,T\}$, the decoder outputs node $v^t$ based on the sequence of node embeddings $H=(h_0,h_1,...,h_N)$ from the encoder and context embedding $C^t$. In MODRL/D-AM, the context embedding $C$ consists of graph embedding $h^{\mathrm{graph}}=\frac{1}{N+1}\sum_{i=0}^N h_i$, the last served node embedding $h^{t-1}$, the depot embedding $h^{\mathrm{depot}}$, the remaining capacity $Q^t$ of the current vehicle, and the current traveling time-point $P^{t}$ of the vehicle. $C^t$ is formally defined as follows:
\begin{equation}
	C^t=[h^{\mathrm{graph}};h^{t-1};h^{\mathrm{depot}};Q^t;P^{t}],
\end{equation}
where ``$[;]$" denotes the concatenation operator.

The decoder consists of a MHA layer and a subsequent layer with just attention weights operating on a masked input sequence. A valid solution under capacity and time windows constraints is promised by mask function. The mask function rules out the nodes that would violate constraints and the nodes that are already served by setting corresponding attention weight to $-\infty$, and thus yield probability of zero, i.e.,
\begin{equation}
	\begin{aligned}
	p_{\boldsymbol{\theta}}(v^t=i|s_{t-1})&=\pi_{\boldsymbol{\theta}}(v^t=i|H,C^{t-1})\\&=\mathrm{attn}(\mathrm{MHA}(C^{t-1},H),\mathrm{mask}(H)),
	\end{aligned}
\end{equation}
where $s_{t-1}$ denotes the partial solution in time step $t-1$.\\
\indent The probability of a complete solution $s$ for instance $G$ can be modeled by the chain rule as follows,
\begin{equation}
	p_{\boldsymbol{\theta}}(s|G)=\prod_{i=1}^T\pi_{\boldsymbol{\theta}}(v^t|H,C^{t-1}).
\end{equation}
\subsection{Training Algorithm: Reinforcement Learning}
\label{sec:ta}
Consider a specific subproblem with its model $\pi_{\boldsymbol{\theta}}$, a well-known policy gradient based REINFORCE algorithm \cite{williams1992simple} is employed to train it.

The gradients of parameters $\boldsymbol{\theta}$ is formulated as follows,
\begin{equation}
	\nabla_{\boldsymbol{\theta}} J(\boldsymbol{\theta}|G)=\mathbb{E}_{s \sim p_{\boldsymbol{\theta}}(\cdot|G)}[(\lambda^\mathrm{T}F(s|G)-b(G))\nabla_{\boldsymbol{\theta}}\mathrm{log}p_{\boldsymbol{\theta}}(s|G) ],
\end{equation}
where $G$ is an input instance, $s$ is a complete solution, $\lambda$ is weight vector and $b(G)$ is a value function to reduce gradient variance. In the training process, the gradients of parameters $\boldsymbol{\theta}$ can be approximated by Monte Carlo sampling as follows,
\begin{equation}
	\nabla_{\boldsymbol{\theta}} J(\boldsymbol{\theta}|G)\approx\frac{1}{B}\sum_{j=1}^B[(\lambda^\mathrm{T}F(s_j|G_j)-b(G_j))\nabla_{\boldsymbol{\theta}}\mathrm{log}p_{\boldsymbol{\theta}}(s_j|G_j) ],
\end{equation}
where $B$ is the batch size. For the training of $M$ subproblem models, the neighborhood-based parameter-transfer strategy \cite{li2020deep} is employed to accelerate the training process.


\begin{figure*} \centering 
	\label{fig}

	\includegraphics[width=2\columnwidth] {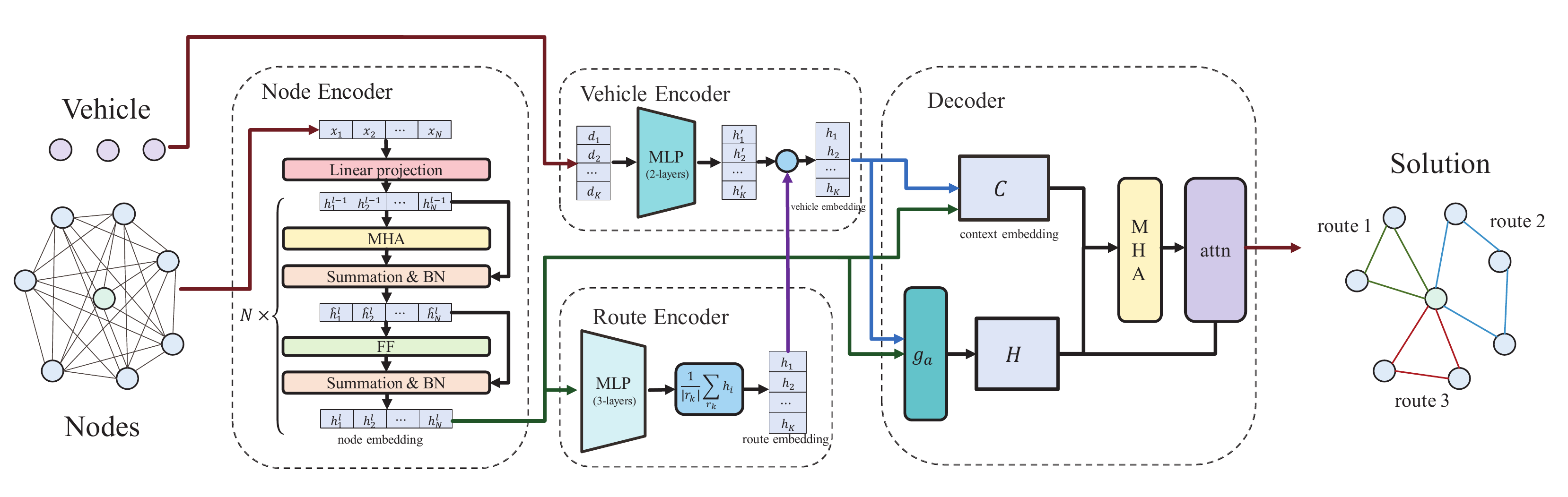}  
	\caption{The overall architecture of our model. In encoder part, the node encoder takes node features as input to compute node embeddings. The route encoder takes node embeddings as input and computes route embeddings. The vehicle encoder takes vehicle features and route embedding as input to compute vehicle embeddings. In the decoder part, the node embeddings and vehicle embeddings that contain comprehensive context information are used to compute context embedding $C$. The node embeddings and vehicle embeddings also used to compute joint combinatorial space $H$. Then, the context embedding $C$ and joint combinatorial space $H$ are used to compute the next visit node.  } 
\end{figure*}

\section{The Proposed Algorithm: MODRL/D-EL}

\subsection{Motivation}
The MODRL/D-AM performs well in simple MOPs (e.g., MO-TSP). In more complicated problems such as MO-VRPTW, the context embedding $C$ in the attention model only contains the information about the vehicle status and the last visited node in current route, while the nodes and vehicles in other routes are not taken into consideration. The limited context information usually leads to sub-optimal decisions. Besides, the diversity of solutions obtained by MODRL/D-AM is not as good as expected \cite{li2020deep}. Therefore, this paper proposes to consider more comprehensive context information in constructing solution, and uses evolutionary learning to fine-tune the parameters of models to improve the diversity of solutions.

\subsection{Model of Subproblem: Attention Model with Comprehensive Context Information}

The proposed model of subproblem is also based on the attention mechanism \cite{kool2019attention}. The node encoder is consistent with the encoder in MODRL/D-AM. To provide comprehensive context information for the decoder to make more reasonable decision in constructing solution, vehicle encoder and route encoder are introduced, and the context embedding is extended as in \cite{falkner2020learning}. The overall architecture of our model is shown in Fig. 1.

Both of the vehicle encoder and route encoder are multi-layer perceptron (MLP) and denoted as $g_v$ and $g_l$, respectively. $g_v$ has 3 layers and $g_l$ has 2 layers. $g_v$ encodes the vehicle feature $d_k$ which consists of vehicle index $k$, the current return distance to depot, the current position (given by coordinates of the last served node) and current time of vehicle $k$. $g_l$ encodes the route feature (given by the nodes served so far). Finally, the embedding of vehicle $k$, denoted as $h_k^{\mathrm{vehicle}}$, is computed by aggregating the information of $g_l$ and $g_v$:
\begin{equation}
	h_k^{\mathrm{vehicle}}=[g_v(d_k);\frac{1}{|r_k|}\sum_{i\in r_k}g_l(h_i^{\mathrm{node}})].
\end{equation}
The new context embedding also contains the information about the entire fleet and the last served nodes of vehicles, defined as follows,
\begin{equation}
	h^{\mathrm{fleet}}=\frac{1}{K}\sum_{k=1}^K h_k^{\mathrm{vehicle}},\quad h^{\mathrm{last}}=\frac{1}{K}\sum_{k=1}^K h_{\mathrm{last}(r_k)}^{\mathrm{node}},
\end{equation}
where $\mathrm{last}(r_k)$ denotes the last node in route $r_k$. Finally, the context embedding is defined as follows,
\begin{equation}
	C=[h^{\mathrm{graph}};h^{\mathrm{fleet}};h^{\mathrm{cur}};h^{\mathrm{depot}};h^{\mathrm{last}}],
\end{equation}
where $h^{\mathrm{cur}}$ denotes the embedding of vehicle in current constructing route, and $h^{depot}$ denotes the embedding of depot.  

To extract features of nodes and vehicles more effectively, the node embedding $h^{\mathrm{node}}$ and vehicle embedding $h^{\mathrm{vehicle}}$ are combined to create a new sequence of node embedding $H$, called joint combinatorial space\cite{falkner2020learning} of all nodes and current vehicle, formally defined as follows,
\begin{equation}
	H=\{g_a(h^{\mathrm{cur}},h_i^{\mathrm{node}})| i \in \mathcal{V}\},
\end{equation}
where $g_a$ is a neural network that computes compatibility between a vehicle and a node, formally defined as follows,

\begin{equation}
	\begin{split}
		g_a(h^{\mathrm{cur}},h_i^{\mathrm{node}};W_1,W_2,&W_3)=W_1h_i^{\mathrm{node}}+W_2h^{\mathrm{cur}}\\
		&+W_3[h^{\mathrm{cur}}\odot h_i^{\mathrm{node}};(h^{\mathrm{cur}})^\mathrm{T}h_i^{\mathrm{node}}].
	\end{split}
\end{equation}
where $\odot$ is the element-wise product.

\begin{algorithm}
	\caption{Evolutionary Learning}
	\label{alg:EL}
	\begin{algorithmic}[1]
		\Require The initial population  $\mathrm{\Pi}=\{\pi_1,\pi_2,...,\pi_M\}$ where $\pi_i$ has parameters $\boldsymbol{\theta}_i$, training batch size $B$, maximum generation $D$
		\Ensure optimized model set $\mathrm{\Pi}^*=\{\pi_1^*,\pi_2^*,...,\pi_M^*\}$

		\For {$i=1,...,M$}
		\State $\mathrm{fitness}_{\pi_i}=\mathrm{Evaluate}(\pi_i)$
		\EndFor
		\For{$g=1,...,D$}
		\For{$i=1,...,M$}
		\State$G_j=\mathrm{RandomInstance}() \  \forall j \in \{1,...,B\}$ 
		\Comment{randomly sample training data}
		\State $\boldsymbol{\epsilon}=\sqrt{\sum_{k=0}^{N}(\sum_{j=1}^B\nabla_{\boldsymbol{\theta}}p(v^{t_j}=k|s_{{t_j}-1},G_j))^2}$
		\State $ \bm{x}\sim \mathcal{N}(\bm{0},\mu \bm{I})$
		\State $\hat {\boldsymbol{\theta}}_i= {\boldsymbol{\theta}}_i+\frac {\bm{x}} {\bm{\epsilon}}$
		\State  generate offspring $\hat \pi_i$ with parameters $\bm{\hat\theta_i}$ and $\mathrm{fitness}_{\hat \pi_i}=\mathrm{Evaluate}(\hat \pi_i)$  
		\EndFor
		\State  sort individuals (policies/models) with non-domination level and crowding distance, and select $M$ individuals for next generation
		\EndFor
		
	\end{algorithmic}
\end{algorithm}

\begin{figure*} \centering 
	\label{fig}  
	\subfigure[] {
		\label{training method}  
		\includegraphics[width=1.2\columnwidth] {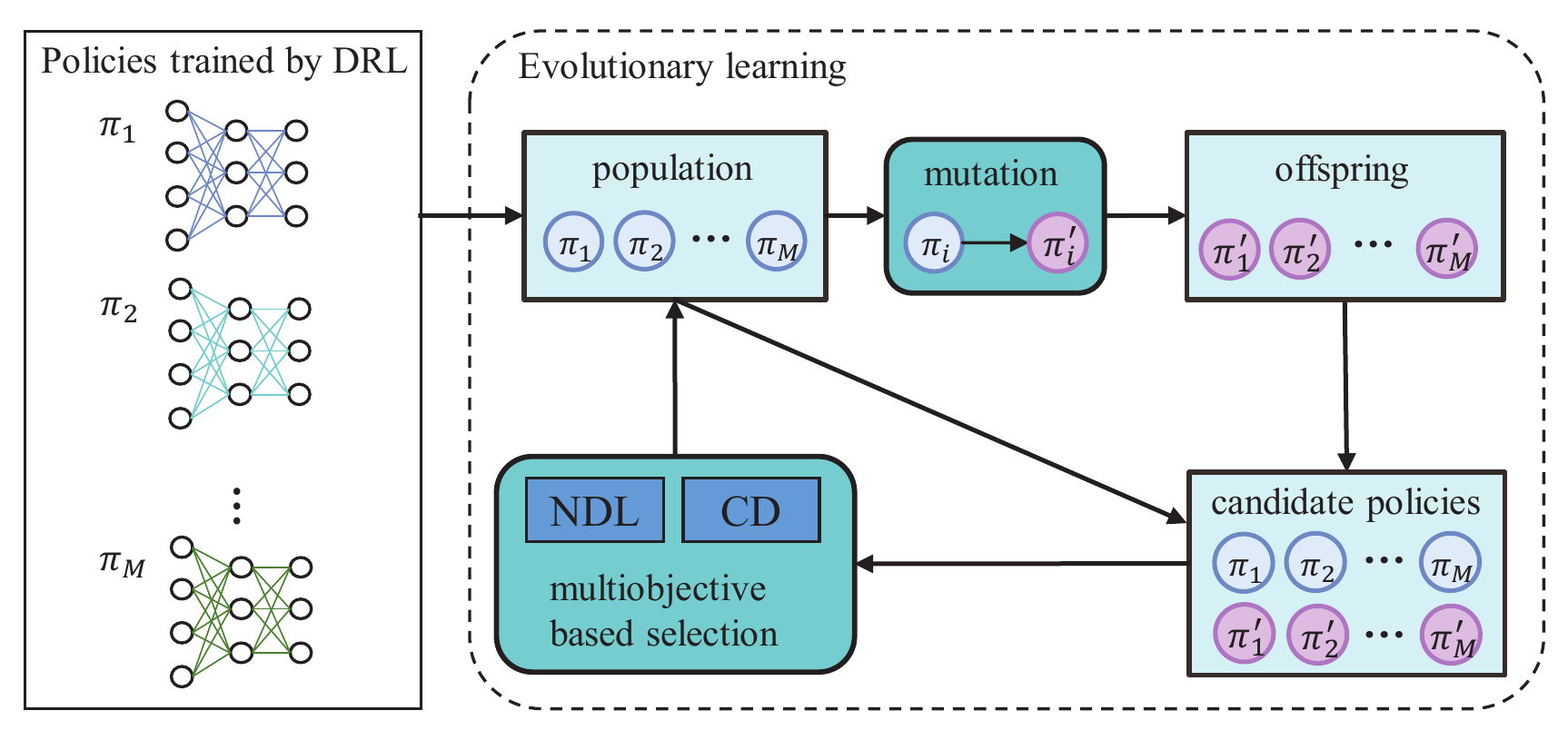}
		
	}    
	\subfigure[] { 
		\label{selection}
		
		\includegraphics[width=0.7\columnwidth]{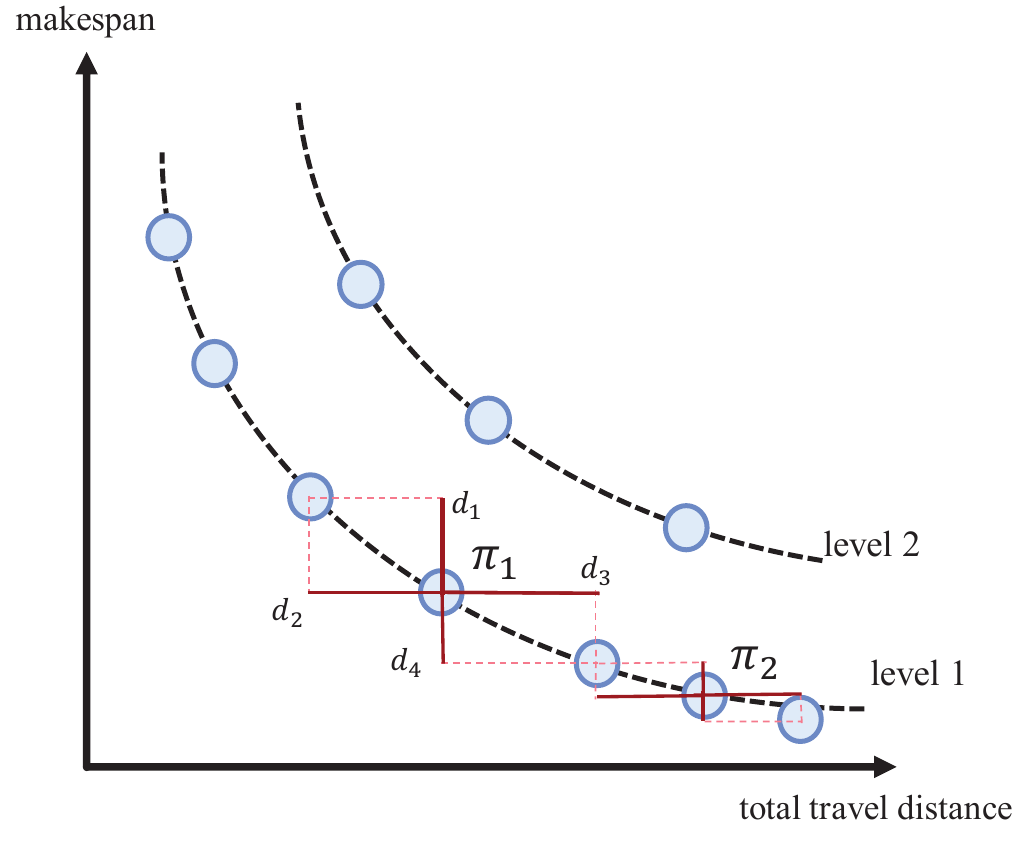}     
	}    
	\caption{The overview of hybrid training algorithm with evolutionary learning.  (a) Hybrid training algorithm with evolutionary learning. (b) Visualization of multiobjective based selection. The non-dominantion level (NDL) is calculated by recursively select non-dominantion policies from the policies set. Then the policies are sorted into different non-dominantion level correspondingly. In the same non-dominantion level, policies are ranked by crowding distance (CD). }    
	\label{fig}     
\end{figure*}

\subsection{Hybrid Training Algorithm with Evolutionary Learning}
The hybrid training algorithm consists of two stages. In the first stage, the models of subproblems are trained by the well-known REINFORCE algorithm \cite{williams1992simple} as described in \textbf{Section \ref{sec:ta}}. The reward of $i$th subproblem is the weighted sum of two objectives as defined in (\ref{objectives}) with weight vector $\lambda_i$.

In the second stage, the models trained by the first stage are taken as the initial population. Then they are evolved by variation operators. At each generation, each individual model generates an offspring by the proximal mutation \cite{bodnar2020proximal}, which takes SM-G-SUM \cite{Lehman2018117} mutation operator to add scaled Gaussian perturbation to the model parameters $\boldsymbol{\theta}$, i.e.,

\begin{equation}
	\boldsymbol{\theta} \leftarrow \boldsymbol{\theta} + \frac{\bm{x}}{\bm{\epsilon}}, \quad \bm{x}\sim \mathcal{N}(\bm{0},\mu \bm{I}),
\end{equation}
where $\mu$ is a mutation magnitude hyperparameter. The SM-G-SUM operator \cite{Lehman2018117} uses the gradient of attention weight of each node in a randomly sampled decoding step over a batch of instances to compute the sensitivity $\boldsymbol{\epsilon}$, formally defined as follows,  
\begin{equation}
	\boldsymbol{\epsilon}=\sqrt{\sum_{k=0}^{N}(\sum_{j=1}^B\nabla_{\boldsymbol{\theta}}p_{\boldsymbol{\theta}}(v_{t,j}=k|s_{t-1,j},G_j))^2},
\end{equation}
where $N$ is the number of nodes, $B$ is the batch size of the training instances $G$, and $t$ is the time step in the decoding process, which is randomly selected from $\{1,2,...,T\}$. $s_{t-1,j}$ is a partial solution at time step $t-1$ for the instance $G_j$, so it gives a random sample state of $G_j$.

The parent and offspring models are evaluated with a randomly sampled MO-VRPTW instance. Specifically, each model takes the instance as input and constructs solution for the instance. Then, a fitness vector (objective vector) is computed for the solution, and the model is marked with the fitness vector. After that, a multi-objective based selection is performed. Specifically, all the models are ranked according to non-domination level and crowding distance of fitness vectors as in NSGA--\uppercase\expandafter{\romannumeral2} \cite{deb2002a}. After ranking, $M$ individuals (policies/models) are reserved. Algorithm \ref{alg:EL} shows the procedure of evolutionary learning. The overview of hybrid training algorithm and multi-objective based selection are shown in Fig. 2.

%
%
%
%

\begin{table*}[h]
	
	\caption{Comparative results (Mean and STD) of HV, $|$NDS$|$ and testing time on three testing sets obtained by MODRL/D-EL, MODRL/D-CCI and MODRL/D-AM, respectively. The highest values of HV are highlighted in \textbf{bold}.}
	\begin{center}
		
		\resizebox{\textwidth}{!}{
			
			\begin{tabular}{|c|ccc|ccc|ccc|}
				\hline
				\multirow{2}*{\diagbox{Instances}{Algorithm}}\multirow{2}{*}{} & \multicolumn{3}{c|}{MODRL/D-EL}   & \multicolumn{3}{c|}{MODRL/D-CCI}  & \multicolumn{3}{c|}{MODRL/D-AM} \\ \cline{2-10} 
				& HV              & $|$NDS$|$     & Time (s) & HV              & $|$NDS$|$     & Time (s) & HV        & $|$NDS$|$       & Time (s)   \\ \hline
				$N=50$ & 730.28 (30.48)          & 14.53 (3.09) & 48.09   & \textbf{731.24 (30.49)} & 12.81 (2.69) & 47.49   & 728.57 (30.79)   & 11.72 (2.56)    & 37.33    \\
				$N=80$ & \textbf{844.45 (37.77)} & 14.90 (2.96) & 68.63   & 843.55 (37.65)         & 11.96 (2.52) & 67.58   & 842.45 (38.29)    & 12.73 (2.58)   & 54.37     \\
				$N=100$ & \textbf{1154.76 (49.34)} & 14.11 (2.83) & 83.07   & 1150.94 (49.57)          & 10.00 (2.29)& 81.02   & 1150.25 (50.36)   & 11.90 (2.44)   & 68.13     \\ \hline
			\end{tabular}
		}
		\label{compare to modrl}
	\end{center}
\end{table*}

\begin{table*}[h]\label{benchmark}
	\caption{The HV values, $|$NDS$|$ and testing time on 11 solomon instances R201-211. The mean values and standard
		deviation (STD) are shown in the bottom. To eliminate the impact of randomness in NSGA-\uppercase\expandafter{\romannumeral2}, MOEA/D and MOGLS, these algorithms are run 5 times independently and the mean values are taken as final results. The highest values of HV are highlighted in \textbf{bold}.}\label{compare to modrl}
	\resizebox{\textwidth}{!}{
		\begin{tabular}{|c|ccc|ccc|ccc||ccc|ccc|ccc|}
			\hline
			\multirow{2}*{\diagbox{Instances}{Algorithm}} & \multicolumn{3}{c|}{MODRL/D-EL}  & \multicolumn{3}{c|}{MODRL/D-CCI} & \multicolumn{3}{c||}{MODRL/D-AM}  & \multicolumn{3}{c|}{NSGA-II} & \multicolumn{3}{c|}{MOEA/D} & \multicolumn{3}{c|}{MOGLS} \\ \cline{2-19} 
			& HV             & $|$NDS$|$ & Time (s) & HV             & $|$NDS$|$ & Time (s) & HV             & $|$NDS$|$ & Time (s) & HV      & $|$NDS$|$   & Time (s)  & HV      & $|$NDS$|$  & Time (s)  & HV     & $|$NDS$|$  & Time (s)  \\ \hline
			R201              & 1296.9          & 14.0  & 63.8    & 1301.3          & 7.0   & 63.2    & \textbf{1309.3} & 8.0   & 53.4    & 1265.1   & 15.0    & 202.3    & 1285.6   & 8.0    & 139.3    & 1236.5  & 7.0    & 175.2    \\
			R202              & \textbf{1308.2} & 12.0  & 60.3    & 1306.0          & 8.0   & 56.4    & 1306.8          & 10.0  & 49.9    & 1275.5   & 15.0    & 205.4    & 1298.1   & 9.0    & 139.3    & 1266.0  & 7.0    & 187.8    \\
			R203              & \textbf{1324.2} & 11.0  & 59.6    & 1299.9          & 9.0   & 56.9    & 1210.0          & 7.0   & 48.3    & 1287.1   & 12.0    & 205.7    & 1311.9   & 15.0   & 138.8    & 1294.5  & 8.0    & 200.3    \\
			R204              & \textbf{1326.6} & 15.0  & 59.2    & 1295.6          & 8.0   & 59.8    & 1197.2          & 8.0   & 49.0    & 1264.9   & 9.0     & 200.8    & 1296.2   & 7.0    & 136.8    & 1302.6  & 10.0   & 200.2    \\
			R205              & 1307.9          & 8.0   & 60.1    & 1313.2          & 10.0  & 56.5    & \textbf{1319.0} & 6.0   & 48.6    & 1276.9   & 17.0    & 206.6    & 1306.4   & 18.0   & 138.5    & 1271.8  & 10.0   & 188.4    \\
			R206              & 1311.2          & 11.0  & 59.8    & \textbf{1316.6} & 8.0   & 55.8    & 1303.6          & 6.0   & 47.9    & 1278.9   & 12.0    & 202.6    & 1310.9   & 10.0   & 139.0    & 1291.8  & 8.0    & 194.7    \\
			R207              & \textbf{1320.7} & 14.0  & 62.3    & 1295.6          & 8.0   & 56.1    & 1199.5          & 3.0   & 47.8    & 1265.5   & 11.0    & 203.1    & 1307.7   & 15.0   & 138.1    & 1301.7  & 13.0   & 200.9    \\
			R208              & \textbf{1332.4} & 17.0  & 59.0    & 1297.9          & 9.0   & 55.9    & 1227.9          & 6.0   & 48.1    & 1243.0   & 5.0     & 202.2    & 1263.9   & 7.0    & 136.1    & 1292.4  & 7.0    & 198.4    \\
			R209              & 1317.2           & 15.0  & 59.9    & 1320.8          & 11.0  & 58.7    & \textbf{1323.5} & 9.0   & 48.5    & 1286.7   & 19.0    & 203.6    & 1315.3   & 16.0   & 138.2    & 1294.5  & 13.0   & 192.0    \\
			R210              & 1318.2          & 14.0  & 59.8    & 1317.6          & 14.0  & 55.6    & \textbf{1322.3} & 9.0   & 50.6    & 1278.0   & 16.0    & 204.9    & 1313.1   & 18.0   & 141.5    & 1293.9  & 14.0   & 190.2    \\
			R211              & 1322.7          & 15.0  & 59.5    & 1325.6         & 12.0  & 55.3    & \textbf{1337.0} & 7.0   & 47.3    & 1263.6   & 7.0     & 203.9    & 1295.8   & 8.0    & 137.3    & 1304.4  & 10.0   & 204.5    \\ \hline
			Mean           & \textbf{1316.9} & 13.3  & 60.3    & 1308.2          & 9.5   & 57.3    & 1277.8          & 7.2   & 49.0    & 1271.4   & 12.5    & 203.7    & 1300.4   & 11.9   & 138.4    & 1286.4  & 9.7    & 193.9    \\
			STD               & 9.7            & 2.4   & 1.4     & 10.4    & 2.0   & 2.3     & 53.5           & 1.8   & 1.7     & 12.1     & 4.1     & 1.7      & 14.5    & 4.3    & 1.4      & 19.5   & 2.5    & 8.0      \\ \hline
		\end{tabular}
	}
\end{table*}

\subsection{The Procedure of MODRL/D-EL}
Algorithm \ref{alg:MODRL/D-EL} shows the procedure of MODRL/D-EL. MODRL/D-EL first  decomposes MO-VRPTW into $M$ single objective subproblems by weighted sum approach. For each subproblem, an attention model with comprehensive context information is trained by reinforcement learning with parameter-transfer strategy \cite{li2020deep}. Then, the trained models are taken as the initial population for evolutionary learning. The evolutionary learning uses proximal mutation \cite{bodnar2020proximal} to generate offsprings, and select individuals (policies/models) according to the ranks of non-domination level and crowding distance as in NSGA--\uppercase\expandafter{\romannumeral2} \cite{deb2002a}. 
\begin{algorithm}
	\caption{MODRL/D-EL}
	\label{alg:MODRL/D-EL}
	\begin{algorithmic}[1]
		\Require The model set of subproblems $\mathrm{\Pi}=\{\pi_1,\pi_2,...,\pi_M\}$, weight vectors $\Lambda=\{\lambda_1,\lambda_2,...,\lambda_M\}$, maximum iteration in evolutionary learning $D$
		\Ensure optimized model set $\mathrm{\Pi^*}=\{\pi_1^*,\pi_2^*,...,\pi_M^*\}$ 
		\State randomly initialize parameters of models in $\mathrm{\Pi}$
		\For{$i=1,...,M$}
		\State train policy $\pi_i$ by DRL as described in \textbf{Section 3.3} for $i$th subproblem 
		\EndFor
		\State take policy set trained by DRL as the initial population for evolutionary learning
		\For{$i=1,...,D$}
		\State evolve individuals (policies/models) as described in \textbf{Algorithm 1}
		\EndFor
	\end{algorithmic}
\end{algorithm}

\section{Experimental Results}
All the experiments are executed on a machine with an Intel Xeon E5-2637 CPU and 16GB memory. All the related network models are trained on a single 1080Ti GPU.
\subsection{Experiment Settings}
\subsubsection{Data Generation}
The vehicles capacity $Q$ is set uniformly to 750, 900 and 1000 for problem size 50, 80 and 100, respectively. The time window of depot is set to $[a_{\mathrm{0}}=0,b_{\mathrm{0}}=10]$, and service duration $w$ of customers is set to 0.1 time unit. The coordinates of nodes, demands and time windows of customers are defined as follows \cite{falkner2020learning}:
\begin{itemize}
	\item {\textbf{Coordinates}}: The coordinates of nodes are sampled uniformly from the interval $[0,1]$.
	\item {\textbf{Demands}}: The demand $q_i$ is sampled from a normal distribution $\hat q_i\sim \mathcal{N}(15,10)$, and truncated its absolute value to integer
	values between 1 and 42, i.e., $q_i'=\mathrm{min}(42,\mathrm{max}(1,\lfloor|\hat q_i|\rfloor))$. 
	Then, $q_i'$ is scaled by the vehicle capacity $Q$, i.e. $q_i=q_i'/Q$. The vehicle capacity $Q$ is also normalized to 1 in training. 
	\item {\textbf{Time Window}}: For customer $v_i$, $h_i=[a_{\mathrm{sample}}=\hat h_i,b_{\mathrm{sample}}=b_0-\hat h_i-w]$ is defined, where $\hat h_i= d_i+0.01$, and $d_i$ is the traveling time from $v_i$ to the depot. Then, the start time $a_i$ is sampled uniformly from $h_i$, and the end time $b_i$ is calculated as $b_i=\mathrm{min}( a_i+3\alpha ,b_{\mathrm{sample}})$, $\alpha=\mathrm{max}(|\hat \alpha|,0.01),\ \hat \alpha\sim\mathcal{N}(0,1).$
\end{itemize}

\begin{figure*}[t] \centering 
	\label{fig}  
	\subfigure[] {
		\label{fig:a}  
		\includegraphics[width=0.9\columnwidth] {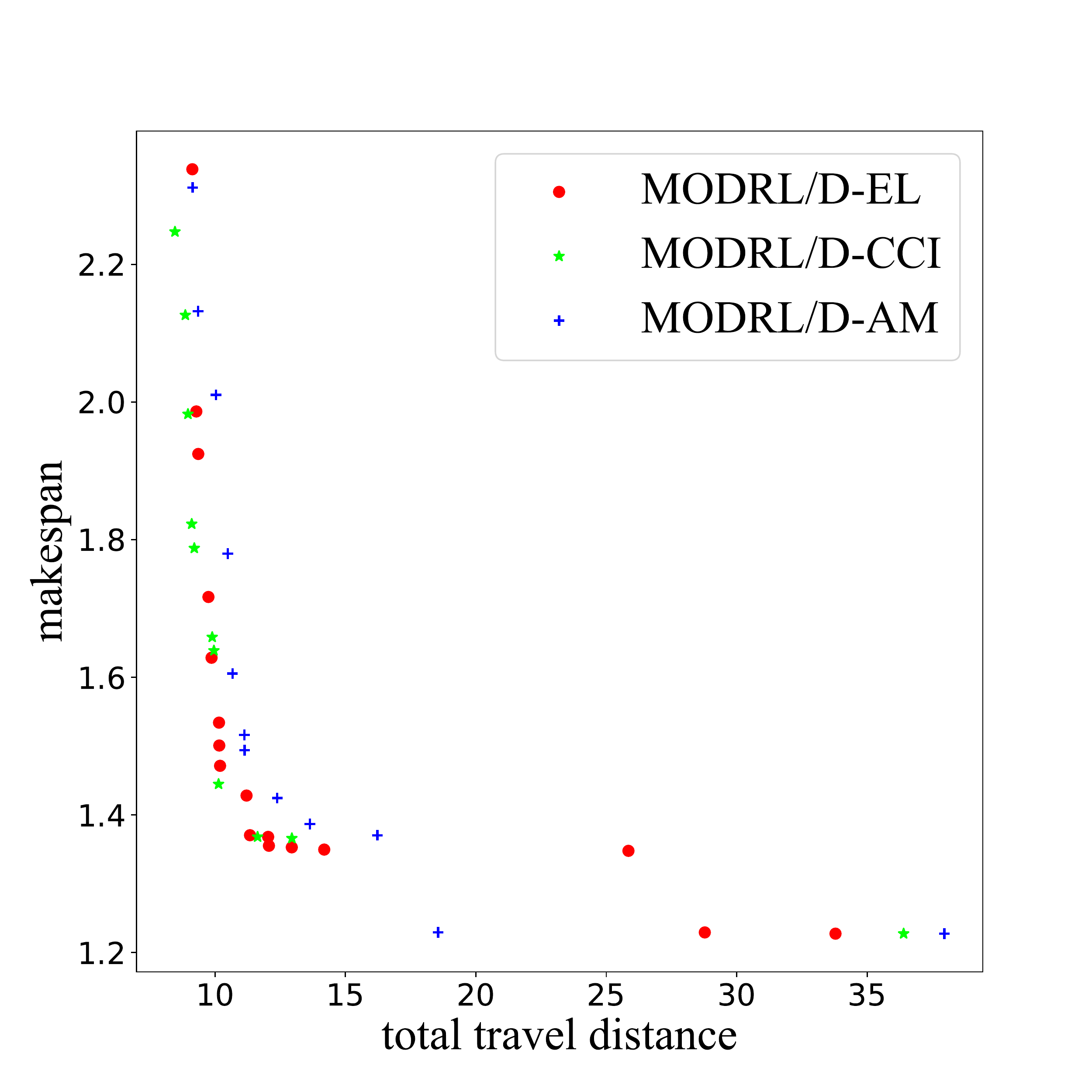}  
	}    
	\subfigure[] { 
		\label{fig:b}
		
		\includegraphics[width=0.9\columnwidth]{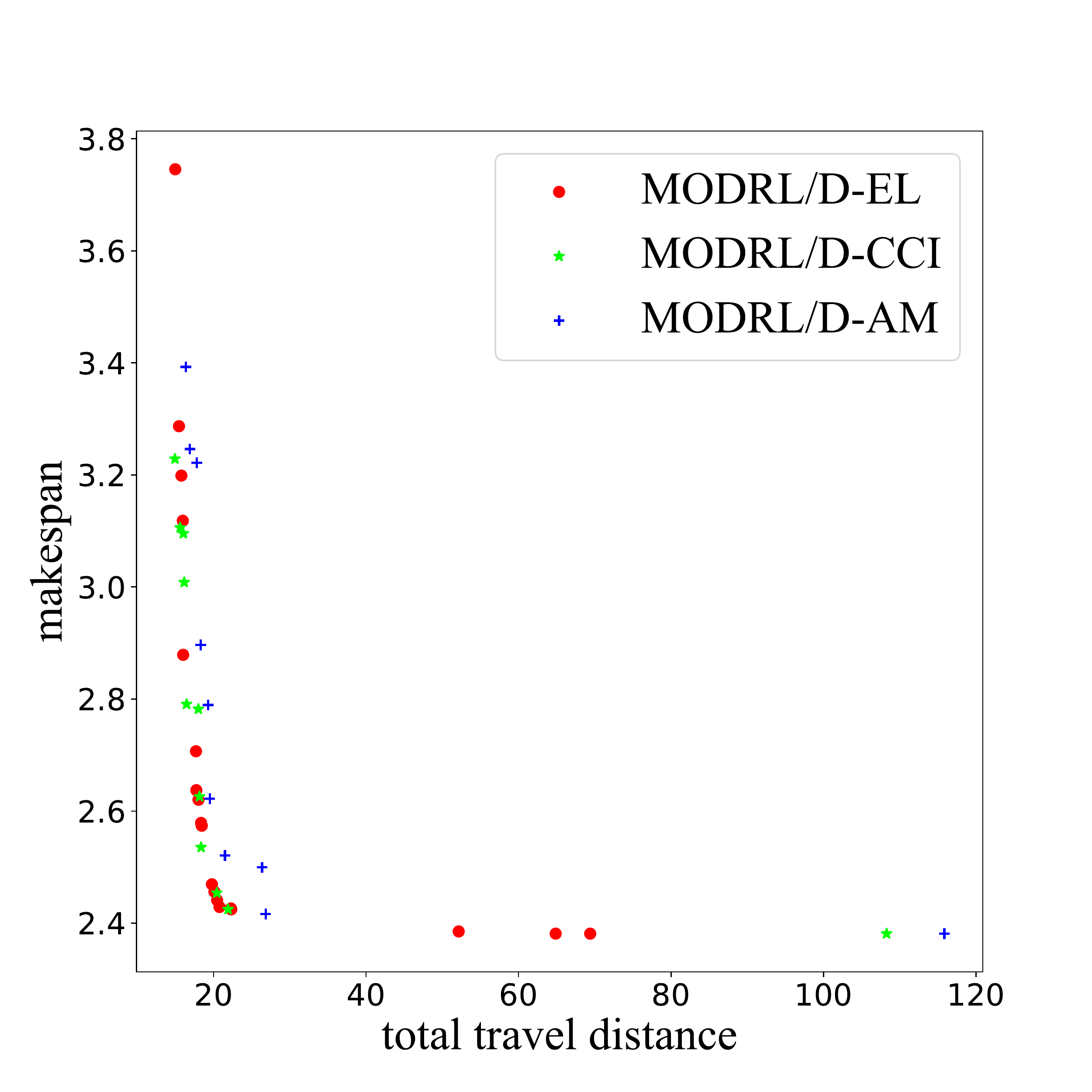}     
	}    
	\caption{The PFs obtained by MODRL/D-EL, MODRL/D-CCI and MODRL/D-AM for random instances with (a) 50 customer nodes, (b) 100 customer nodes. }     
	\label{fig}     
\end{figure*}

\subsubsection{Training Details}
In deep reinforcement learning, the subproblem number $M$ is set to 100. The weight vectors for subproblems are uniformly spread from 0 to 1, i.e. $\lambda_1=(0,1),\lambda_2=(0.01,0.99),...,\lambda_{100}=(0.99,0.01)$.  The value function in REINFORCE algorithm\cite{williams1992simple} is the greedy rollout baseline proposed in \cite{kool2019attention}. Adam \cite{kingma2015adam} is used as optimizer, and the learning rate is set to $10^{-4}$ constantly. 
The number of heads in MHA is set to 8. 
The dimension $d_{\mathrm{emb}}$ of $h^{\mathrm{node}}$ and $h^{\mathrm{vehicle}}$ are set to 128. The hidden dimension in FF layer is 256. The hidden dimension of $g_v$ and $g_l$ are set to 64. 

Based on the neighborhood-based parameter-transfer strategy \cite{li2020deep}, the first model is trained for 10 epochs, each epoch involving 512,000 instances. The remaining 99 models are only trained 1 epoch; each epoch contains 256,000 instances. The training batch size in DRL is set to 128.  

In evolutionary learning, the hyperparameter $\mu$ in proximal mutation is set to $0.01$. The training batch size in evolutionary learning is set to 64. The maximum generation $D$ is set to 30.

The training time for our models is about 20 hours.

\subsubsection{Baselines}

Two learning-based algorithms are introduced as competitors to evaluate the performance of MODRL/D-EL. The first one is MODRL/D-AM \cite{wu2019modrl}, and the second one is MODRL/D-CCI, which extends MODRL/D-AM by integrating the comprehensive context information. MODRL/D-CCI is compared with MODRL/D-AM to evaluate the contribution of CCI. MODRL/D-EL is compared with MODRL/D-CCI to evaluate the contribution of evolutionary learning.

MODRL/D-EL is also compared with three classical iterative based algorithms: NSGA-\uppercase\expandafter{\romannumeral2}, MOEA/D and MOGLS. For NSGA-\uppercase\expandafter{\romannumeral2} and MOEA/D, the number of iterations is set to 20000. The population size is set to 100. For MOGLS, the size of the temporary population and the number of the initial solutions are consistent with the setting in \cite{jaszkiewicz2002genetic}. The local search uses 2-opt operator, which is terminated when 100 iterations are reached. The number of iterations in MOGLS is set to 4000 for a similar testing time with NSGA-\uppercase\expandafter{\romannumeral2}.

\subsection{Results and Discussions}

All the attention models used in learning-based algorithms are trained with the instances of size 50. The average performance over a batch of instances can evaluate the algorithms more accurately. The learning-based algorithms learn from the training data and thus, they can efficiently construct solutions for a batch of instances from the same distribution. The iterative-based algorithms do not require the tested instances to follow some distribution. However, the iteration process needs to be re-performed for every run. Hence, a tremendous amount of time is required to solve a batch of instances.

The experiments consist of two parts. For the first part, to compare MODRL/D-EL with other learning-based algorithms, three testing sets of problem size 50, 80 and 100 are respectively sampled from the same distribution as training data. Each testing set contains 1024 instances. The iterative-based algorithms are not tested in this part, because these algorithms take a tremendous amount of time to solve 1024 instances. For the second part, to compare MODRL/D-EL with iterative-based algorithms, 11 instances (R201--R211) from Solomon benchmark \cite{castro-gutierrez2011nature} are used as testing instances. Hypervolume (HV) and the number of solutions in non-dominant set ($|$NDS$|$) are taken as performance indicators.  The reference point for calculating HV are set to (100,10), (120,10) and (160,10) for problem size 50, 80 and 100 respectively.

Table II shows the experimental results on the three testing sets. In comparisons between MODRL/D-CCI and MODRL/D-AM, MODRL/D-CCI achieves better performance in term of HV on all the three testing sets, which suggests that comprehensive context information is helpful to make more reasonable decision. In comparisons between MODRL/D-EL and MODRL/D-CCI, MODRL/D-EL achieves better performance in term of HV on the testing sets of problem size 80 and 100. In addition, MODRL/D-EL obtains more non-dominant solutions on all the three testing sets. The fact suggests that evolutionary learning can fine-tune the parameters of models to improve the quality of solutions. The testing time of MODRL/D-EL is a little longer than that of MODRL/D-AM. This is because processing comprehensive context information requires more testing time than processing simple context information. To further visualize the final solutions found by different approaches, Fig. 2 shows the approximate Pareto fronts of two instances with different problem size. As shown in the figure, MODRL/D-EL has a better performance in both convergence and diversity.

Table III reports the experimental results on 11 benchmark instances. In comparisons between MODRL/D-CCI and MODRL/D-AM, MODRL/D-CCI achieves better mean values in terms of HV and $|$NDS$|$, and MODRL/D-CCI has a much more stable performance. In comparisons between MODRL/D-EL and MODRL/D-CCI, MODRL/D-EL achieves better mean value in term of HV, and MODRL/D-EL obtains much more non-dominant solutions in 10 out of 11 instances. In comparisons between MODRL/D-EL and iterative-based algorithms, MODRL/D-EL is much faster in solving all the instances and gets better mean values. Besides, even though MODRL/D-EL is trained with instances of problem size 50, it still performs better than iterative-based algorithms on the benchmark instances of problem size 100, which suggests that MODRL/D-EL has a good generalization ability to tackle the problem with different sizes.

\section{Conclusion}
This paper proposes a multiobjective deep reinforcement learning and evolutionary learning algorithm for MOPs with complex constraints. In the proposed algorithm, the attention model constructs solution by using more comprehensive context information of MOP, which is helpful in making reasonable decision. The evolutionary learning is employed in the proposed algorithm to fine-tune the parameters of models to achieve better performance. The proposed algorithm is compared with other learning-based and iterative-based algorithms for MO-VRPTW. The experimental results suggest that the proposed algorithm has a better performance in both convergence and diversity.

With respect to the future studies, the evolutionary learning in this paper only considers mutation operator. A crossover operator may generate offspring models that construct solutions with higher diversity and convergence. Besides, the MO-VRPTW instances in this paper are symmetrical. How to efficiently deal with asymmetric instances is also worth further studying.

%


\bibliography{./ref}

\bibliographystyle{./bibliography/IEEEtran}

\end{document}